\documentclass[11pt]{article}

\usepackage[final]{acl}

\usepackage{times}
\usepackage{latexsym}

\usepackage[T1]{fontenc}

\usepackage[utf8]{inputenc}

\usepackage{microtype}

\usepackage{inconsolata}

\usepackage{graphicx}
\usepackage{tikz}
\usetikzlibrary{positioning,calc,fit,backgrounds,matrix}
\usepackage{subcaption}
\usepackage{booktabs}
\usepackage{amsmath}
\usepackage{amssymb}
\usepackage{bm}
\usepackage{glossaries}
\usepackage{stfloats}

\newcommand{\coderepo}{\href{https://github.com/wullli/softirr}{\nolinkurl{github.com/wullli/softirr}}}

\title{Consensus Measures for Unstructured Biomedical Text Annotations}

\author{
  Pascal Wullschleger$^{1,3}$,
  Christian Kreis$^2$,
  Martin A. Walter$^2$\\
  \textbf{Jennifer Foster$^1$},
  \textbf{Marc Pouly$^3$} \\\\
  $^1$ Hamilton Institute, Maynooth University, Ireland\\
  $^2$ University of Lucerne, Switzerland\\
  $^3$ Lucerne School of Computer Science and IT, Switzerland\\\\
  \texttt{pascal.wullschleger@hslu.ch}
}

\defcitealias{meddra2026}{ICH, 2025}

\newacronym{LLM}{LLM}{large language model}
\newacronym{NLP}{NLP}{natural language processing}
\newacronym{IRR}{IRR}{inter-rater reliability}
\newacronym{NLI}{NLI}{natural language inference}
\newacronym{LCS}{LCS}{longest common subsequence}
\newacronym{ICD}{ICD}{International Classification of Diseases}
\newacronym{MeSH}{MeSH}{Medical Subject Headings}
\newacronym{MedDRA}{MedDRA}{Medical Dictionary for Regulatory Activities}
\newacronym{MAE}{MAE}{mean absolute error}
\newacronym{MASI}{MASI}{Measuring Agreement on Set-valued Items} 
\newacronym{ECE}{ECE}{expected calibration error}
\newacronym{KS}{KS}{Kolmogorov--Smirnov}
\newacronym{REFLACX}{REFLACX}{Reports and Eye-Tracking Data for Localization of Abnormalities in Chest X-rays}
\newacronym{NCIt}{NCIt}{National Cancer Institute Thesaurus}

\begin{document}
\maketitle
\begin{abstract}
Biomedical literature is increasingly mined for knowledge beyond the questions it was written to answer.
Because the target concepts are not known in advance, annotators prefer open-ended labels, whose agreement is hard to
quantify.
We study \emph{soft inter-rater reliability} for annotators providing unstructured texts for biomedical annotation tasks. 
Synthetic experiments show that soft reliability can be quantified using a variety of semantic equivalence
measures, and that the choice of measure affects failure modes of the estimation. Embeddings are scalable, 
but limited when differentiating similar but distinct concepts. Large
language models are promising, but limited by scalability for estimating agreement by chance.
Finally, measures based on natural language inference can be a compromise, but application to long form annotations requires care.%
{\renewcommand{\thefootnote}{*}\footnote{Code available at \coderepo.}\addtocounter{footnote}{-1}}
\end{abstract}

\section{Introduction}

\begin{figure}[t]
    \centering
	\def\pematrixwidth{1.0\linewidth}
\providecommand{\pematrixwidth}{\linewidth}
\resizebox{\pematrixwidth}{!}{%
\begin{tikzpicture}[
  cell/.style={
    draw=gray!45, minimum width=1.45cm, minimum height=0.35cm,
    align=center, font=\scriptsize, inner sep=1pt
  },
  diagcell/.style={cell, draw=black, line width=1pt},
  rowlbl/.style={font=\scriptsize\itshape, anchor=east},
  collbl/.style={font=\scriptsize\itshape, anchor=south, align=center, text width=1.4cm, inner sep=1pt},
  header/.style={font=\scriptsize\itshape},
  panel/.style={draw=black, dotted, thick, rounded corners=2pt},
  score/.style={font=\scriptsize},
]

\def\cw{1.5}
\def\ch{0.40}

\def\xmid{\cw}                  
\def\ymid{-\ch}                 
\def\xlbl{-2.35}                
\def\xleft{-2.55}               
\def\xright{2*\cw+0.75}         
\def\xpanelmid{0.4}             
\def\ytop{1.10}                 
\def\ybot{-2*\ch-0.70}          

\newcommand{\simmatrix}[5]{%
  \begin{scope}[shift={(#2,#3)}]
    \foreach \val [count=\k from 0] in {#4} {
      \pgfmathtruncatemacro{\r}{div(\k,3)}
      \pgfmathtruncatemacro{\c}{mod(\k,3)}
      \pgfmathsetmacro{\xx}{\c*\cw}
      \pgfmathsetmacro{\yy}{-\r*\ch}
      \foreach \shade [count=\m from 0] in {#5} {
        \ifnum\m=\k
          \ifnum\r=\c
            \node[diagcell, fill=blue!\shade] (#1\r\c) at (\xx,\yy) {\val};
          \else
            \node[cell, fill=blue!\shade] (#1\r\c) at (\xx,\yy) {\val};
          \fi
        \fi
      }
    }
  \end{scope}
}

\newcommand{\panelframe}[5]{%
  \begin{scope}[shift={(#1,#2)}]
    \node[rowlbl] at (-0.75,0)       {Fatigue};
    \node[rowlbl] at (-0.75,-\ch)    {Weakness};
    \node[rowlbl] at (-0.75,-2*\ch)  {Nausea};
    \node[collbl] at (0,0.30)         {Fatigue};
    \node[collbl] at (\cw,0.30)       {Dry eyes};
    \node[collbl] at (2*\cw,0.30)     {Vomiting};
    \node[header, rotate=90] at (\xlbl,\ymid) {Annotator 1};
    \node[header] at (\xmid,0.90) {Annotator 2};
    \node[score] at (\xpanelmid,-2*\ch-0.58) {#4};
    \node[panel, fit={(\xleft,\ytop) (\xright,\ybot)}] (box#5) {};
    \node[anchor=north west, font=\scriptsize\bfseries, xshift=4pt, yshift=-4pt]
      at (box#5.north west) {#3};
  \end{scope}
}

\simmatrix{L}{0}{0}{1,0,0,0,0,0,0,0,0}{55,3,3,3,3,3,3,3,3}
\panelframe{0}{0}{Traditional \acrshort{IRR}}{$p_o = 0.33$\quad $p_e = 0.11$\quad $\kappa = 0.25$}{Lbox}

\simmatrix{R}{0}{-3.00}{1.00,0.18,0.10,0.55,0.13,0.08,0.12,0.07,0.72}%
  {55,12,7,32,9,6,8,5,42}

\panelframe{0}{-3.00}{Soft \acrshort{IRR}}{$p_o^s = 0.62$\quad $p_e^s = 0.33$\quad $\kappa^s = 0.43$}{Rbox}

\end{tikzpicture}%
}
	\caption{Toy example highlighting the difference between traditional and soft \acrshort{IRR} measures.
	Row and column labels indicate free-text annotations of annotators 1 and 2, respectively.}
\label{fig:agreement_toy_example}
\end{figure}

Evidence synthesis in biomedical research has traditionally focused on answering clinical questions by
combining results from studies investigating predefined variables of interest. Publications are used
largely for the purpose for which they were originally designed \cite{fixed_annotations}.
There, the annotation space is known in advance, and measuring \gls{IRR} on the resulting codified
annotations \cite{cohens_kappa,scotts_pi,fleiss_kappa} is standard procedure to enable insights into
their reliability and quality \cite{cl_irr, slr_irr}.
Such annotations rely on standardised terminologies, which annotators either apply directly or map to
retrospectively \cite{coding_interviews,disease_corpus}.

However, biomedical literature is increasingly repurposed as a source of reusable knowledge, such as
molecular targets, imaging biomarkers or treatment toxicities, beyond the questions originally studied by
the authors \cite{repurpose1,repurpose2,repurpose3}, so that the relevant concepts are not fully known
a priori.
In such exploratory settings, open-ended annotations are preferred, since closed settings require prior
knowledge and can bias annotators \cite{open_closed}, standardising too early risks losing uncommon or
undescribed observations \cite{chan2021utility}, and defining codebooks is time consuming and expensive
\cite{coding_cost,inductive_coding}.
Exploratory annotation therefore begins with unrestricted natural-language descriptions, later
harmonised to controlled vocabularies such as the \gls{MedDRA} \citepalias{meddra2026}, \gls{MeSH}
\cite{mesh2026} or the \gls{ICD} \cite{who_icd11,icdpaper} once the range of extracted concepts is better
understood.

This creates a methodological challenge for assessing agreement, since conventional measures
assume discrete categories and define agreement through identity.
This is particularly problematic as literature mining increasingly combines human and automated extraction
\cite{repurpose1,repurpose2,kappa4eval,slr_irr}, where a human expert and an automated system might 
describe information in different terms.
The objective is not that an automated system reproduces the exact words or codes selected by a human
annotator, but that both extractors recover the same underlying biomedical information.
We term the assessment of \gls{IRR} on open-ended annotations \emph{soft \gls{IRR}}, since it requires us
to define how equivalent two annotations are on a continuous scale. 
Figure~\ref{fig:agreement_toy_example} illustrates this for two annotators and three items: the observed agreement
$p_o^s$ is the mean diagonal, and the expected chance agreement $p_e^s$ is the mean of all values in the table,
since all annotations have $\frac{1}{3}$ as an empirical label frequency per rater.
This is Cohen's weighted $\kappa$ with model-derived weights. As open-ended text we consider annotations in the form of phrases, keywords,
sentences or paragraphs that are free-text findings of annotators regarding a dataset example (document, image),
but not necessarily verbatim extractions as in spans.

Existing approaches to soft \gls{IRR} are mostly embedding-based \cite{related_softirr} or use fuzzy
matching \cite{masi, related_softirr}, and \gls{IRR} has been used to evaluate annotations generated by
\glspl{LLM} \cite{kappa4eval}, but not for open-ended ones.
To our knowledge, different measures, such as embeddings, \gls{NLI} models or \glspl{LLM},
for measuring semantic equivalence in soft \gls{IRR} have not been systematically compared in terms of performance and
failure modes.

Our contributions are:\ (1) We compare different approaches to soft agreement on biomedical annotations
using controlled synthetic experiments involving standardised terminologies as well as real-world biomedical
datasets, and gain insights into their failure modes and limitations.
(2) We extend our experiments to comparing \emph{sets} of natural language annotations and propose soft
\gls{IRR} approaches for such settings, comparing different methods for solving the assignment problem
between two sets.
(3) We introduce previously unexplored ways to calculate soft \gls{IRR} using \gls{NLI} models inspired
by paraphrase detection \cite{mutual_entailment1, mutual_entailment2}.

\section{Preliminaries}


\subsection{Inter-Rater Reliability Measures}
\paragraph{Average Agreement.}
Average agreement refers to the ratio of annotated examples the annotators agree on
(Equation~\ref{eq:avg_agreement}). Here $l_i^{(1)}$ and $l_i^{(2)}$ are the labels of annotators $1$ and $2$ for example
$i$, and $\mathbf{1}[\cdot]$ is the indicator function that returns 1 if the condition is true and 0 otherwise. In
$\kappa$-statistics, this is called observed agreement, denoted $p_o$.

\begin{equation}
\label{eq:avg_agreement}
p_o := \frac{1}{N} \sum_{i=1}^{N} \mathbf{1}\bigl[l_i^{(1)} = l_i^{(2)}\bigr]
\end{equation}

\paragraph{\texorpdfstring{Cohen's $\bm{\kappa}$}{Cohen's κ}.}
Cohen's $\kappa$ \cite{cohens_kappa} measures \gls{IRR} by the observed agreement $p_o$ corrected for the probability of
agreement by chance $p_e$ (Equation~\ref{eq:kappa}).
It is generally considered more robust than simple average agreement \cite{cl_irr}.
The probability of agreement expected by chance $p_e$ (Equation~\ref{eq:pe}) is the sum over labels $k$ of the product
of the empirical label
frequencies $\hat{p}_k^{(1)}$ and $\hat{p}_k^{(2)}$, where $\hat{p}_k^{(r)}$ is the ratio of examples with label $k$
to the total number of examples $N$ for annotator $r$ (Equation~\ref{eq:pe2}).

\begin{align}
\kappa &:= \frac{p_o - p_e}{1 - p_e} \label{eq:kappa}\\
p_e &:= \sum_{k} \hat{p}_k^{(1)} \cdot \hat{p}_k^{(2)} \label{eq:pe}\\
\hat{p}_k^{(r)} &= \frac{1}{N}\sum_{i=1}^{N} \mathbf{1}\bigl[l_i^{(r)} = k\bigr] \label{eq:pe2}
\end{align}

\paragraph{\texorpdfstring{Fleiss' $\bm{\kappa}$}{Fleiss' κ}.}
Fleiss' $\kappa$ \cite{fleiss_kappa} was originally designed as the extension of Cohen's $\kappa$ to the multi-rater
case.
However, it is rather the extension of Scott's $\pi$ \cite{scotts_pi},
since it assumes that all raters share an empirical label distribution \cite{fleiss_vs_scotts}.
Its definition is given in Appendix~\S\ref{sec:appendix_fleiss}.

\subsection{Equivalence Estimators}
When comparing two annotations $a$ and $b$, we argue that the most natural way to measure agreement is
to estimate the probability of semantic equivalence $P(a \equiv b)$ of the two annotations.
In the subsequent paragraphs, we describe different approaches 
to estimate this probability. 

\paragraph{Set-Overlap.} The simplest equivalence estimates are set overlaps between tokenised strings. We study two
such common methods. Firstly, the Jaccard coefficient \cite{jaccard} is defined by the cardinality of the intersection
over the cardinality of the union over two sets of tokens (Appendix Equation~\ref{eq:jaccard}). The \gls{MASI}
equivalence is a weighted version of the Jaccard equivalence that penalises non-subset relations between two sets
(Appendix Equation~\ref{eq:masi}).

\paragraph{Edit-Distance.}
The edit-based equivalence is simply one minus the normalised edit distance. We consider the Levenshtein
\cite{Levenshtein1965BinaryCC} and Indel \cite{indel} equivalences, as well as the \gls{LCS} equivalence
(Appendix Equations~\ref{eq:norm_esim}--\ref{eq:lcs}).

\paragraph{Mutual Entailment.} \gls{NLI} aims to model semantic entailment between premise ($p$) and hypothesis ($h$)
strings, which we denote as $p \Rightarrow h$. This means that, given a premise, the hypothesis is
either entailed (must be true), contradicted (must be false) or neutral. We estimate the equivalence of two strings $a$ and $b$ based on
\gls{NLI} by mutual entailment (Equation~\ref{eq:eqv_nli}) with the geometric mean of the two directions. Mutual
entailment approaches have been used in machine translation evaluation \cite{mt_entail} and paraphrase
detection \cite{mutual_entailment1, mutual_entailment2}, since two texts that entail each other can be considered
semantically equivalent.

\begin{equation}
\label{eq:eqv_nli}
\hat{P}_\text{NLI}(a \equiv b) = \sqrt{\hat{P}(a \Rightarrow b) \cdot \hat{P}(b \Rightarrow a)}
\end{equation}

\paragraph{Semantic Similarity.}
The embedding-based equivalence is the cosine similarity between two string encodings (vectors) given by, e.g.,
a sentence embedding model. While such methods are widely used, e.g., for retrieval \cite{mteb},
they do not cross-encode the two strings, making them unable to incorporate information about the combination of the two
strings into their prediction. However, since they do not cross-encode,
we only need one forward pass per item instead of per item-pair.
They are commonly trained with contrastive, triplet or ranking losses \cite{simcse,sbert},
which optimise for similarity and do not imply equivalence, e.g.,
two things can be similar without having the same meaning.

\paragraph{\acrshort{LLM}-as-a-Judge.}
We define the \gls{LLM}-as-a-judge equivalence as the mutual entailment as predicted by an \gls{LLM}
(Equation~\ref{eq:eqv_llm}).
We propose two variants: \emph{log-prob} and \emph{verbalised}.
The log-prob variant uses the next-token probabilities returned by the model to compute the probability of entailment in
both directions.\ Here, $\ell_\text{yes}(a, b)$ denotes the log-probability of the model predicting \emph{yes} given a
prompt that 
asks whether $a$ entails $b$ (Equation~\ref{eq:l_t}). Subtracting the log-probability of \emph{no} from this gives the
log-odds, which are then passed through a
sigmoid to obtain a probability in $[0, 1]$ (Equation~\ref{eq:ent_dir}). The verbalised variant uses a JSON output to
return $P_\text{LM}(t \mid \text{prompt}(a,b))$ as a string.

\begin{align}
	\hat{P}(a \Rightarrow b) &= \sigma\!\left(\ell_\text{yes}(a,b) - \ell_\text{no}(a,b)\right) \label{eq:ent_dir} \\
\ell_t(a,b) &= \log P_\text{LM}(t \mid \text{prompt}(a,b)) \label{eq:l_t}\\
\hat{P}_\text{LLM}(a \equiv b) &= \sqrt{\hat{P}(a \Rightarrow b) \cdot \hat{P}(b \Rightarrow a)}
\label{eq:eqv_llm}
\end{align}

\subsection{The Assignment Problem}
An assignment problem arises when we want to compare sets of unstructured text annotations and involves finding the best
match between two sets of nodes in a weighted bipartite graph \cite{hungarian}. 

Given sets $A$ and $B$ with $|A| \le |B|$ (otherwise we swap them) and a pairwise equivalence measure $\hat{P}$, an
assignment is an injection $\phi \colon A \hookrightarrow B$ that associates each element of $A$ with a distinct element
of $B$. The optimal assignment $\phi^*$ maximises the total estimated equivalence across all matched pairs
(Equation~\ref{eq:hungarian}). The two sets need not have equal size, in which case only $|A|$ elements are matched
and the surplus of $B$ remains unmatched. The Hungarian algorithm solves the assignment problem exactly with time
complexity $O(\max(|A|,|B|)^3)$.
It has been used frequently throughout machine learning for matching sets \cite{entity_alignment_hungarian,detr}.
We assume the complexity is rarely problematic because the number of annotations per dataset sample is usually
limited. Otherwise, the annotation task would become costly. 

\begin{equation}
\label{eq:hungarian}
\phi^* = \arg\max_{\phi \colon A \hookrightarrow B} \sum_{a \in A} \hat{P}\bigl(a \equiv \phi(a)\bigr)
\end{equation}

\paragraph{Greedy matching.} Alternatively, we can use greedy matching by iterating over the first set and always
choosing the best match in the second set. This has a time complexity of $O(|A||B|)$,
but does not guarantee an optimal solution.
Figure~\ref{fig:assignment_toy_example} illustrates how a greedy approach can yield a
suboptimal result compared to the Hungarian algorithm.

\begin{figure}[t]
  \centering
  \resizebox{\linewidth}{!}{%
  \begin{tikzpicture}[
    annot/.style={
      draw, rounded corners=4pt,
      minimum width=3.0cm, minimum height=0.55cm,
      align=center, font=\small\itshape, inner sep=5pt
    },
    agreen/.style={annot, fill=green!20,  draw=green!60!black},
    ablue/.style ={annot, fill=blue!12,   draw=blue!60!black},
    ared/.style  ={annot, fill=red!18,    draw=red!65!black},
    dashedge/.style={gray!55, dashed, thin},
    lbl/.style={font=\scriptsize\itshape, inner sep=2pt,
                fill=white, fill opacity=1, text opacity=1, text=gray!70},
  ]
  \node[agreen] (L1) at (0,  0.0) {Fatigue};
  \node[ablue]  (L2) at (0, -1.45) {Weakness};
  \node[ared]   (L3) at (0, -2.90) {Nausea};
  \node[ablue]  (R1) at (6.0,  0.0) {Asthenia (grade 2)};
  \node[agreen] (R2) at (6.0, -1.45) {Lethargy};
  \node[ared]   (R3) at (6.0, -2.90) {Nausea / vomiting};
  \node[font=\small] at (0,    0.70) {Annotator 1};
  \node[font=\small] at (6.0,  0.70) {Annotator 2};
  \draw[dashedge] (L1.east) -- (R1.west);
  \draw[dashedge] (L1.east) -- (R3.west);
  \draw[dashedge] (L2.east) -- (R2.west);
  \draw[dashedge] (L2.east) -- (R3.west);
  \draw[dashedge] (L3.east) -- (R1.west);
  \draw[dashedge] (L3.east) -- (R2.west);
  \draw[blue!60!black,  thick] (L2.east) -- (R1.west);
  \draw[green!60!black, thick] (L1.east) -- (R2.west);
  \draw[red!65!black,   thick] (L3.east) -- (R3.west);
  \node[lbl, text=gray!70]        at ($(L1.east)!0.50!(R1.west) + (0,  0.22)$) {0.90};
  \node[lbl, text=blue!60!black]  at ($(L2.east)!0.20!(R1.west) + (0, -0.22)$) {0.85};
  \node[lbl, text=green!60!black] at ($(L1.east)!0.80!(R2.west) + (0,  0.22)$) {0.88};
  \node[lbl, text=red!65!black]   at ($(L3.east)!0.50!(R3.west) + (0,  0.22)$) {0.78};
  \node[lbl] at ($(L1.east)!0.10!(R3.west) + (0, -0.28)$) {0.10};
  \node[lbl] at ($(L2.east)!0.65!(R2.west) + (0, -0.22)$) {0.72};
  \node[lbl] at ($(L2.east)!0.90!(R3.west) + (0,  0.22)$) {0.08};
  \node[lbl] at ($(L3.east)!0.85!(R1.west) + (-0.30, -0.22)$) {0.15};
  \node[lbl] at ($(L3.east)!0.18!(R2.west) + (0,  0.22)$) {0.20};
  \end{tikzpicture}}
  \caption{Toy example of suboptimal greedy assignment for two sets of annotations.
  Edge weights are estimated equivalences $\hat{P}(a \equiv b)$.
  Coloured edges show the Hungarian optimal matching (mean equivalence 0.84).
  Greedy would first match Fatigue to Asthenia (0.90, dashed), resulting in a mean equivalence of 0.80.}
  \label{fig:assignment_toy_example}
\end{figure}

\section{Related Work}
\label{sec:related_work}

Cohen's work on weighted $\kappa$ \cite{cohen_weighted} introduced a notion of distance between rated categories, such 
that one can specify that some disagreements are not always of equal importance.
This is perhaps the first notion of what we call soft \gls{IRR}.
However, here categories are still discrete and distances are defined a priori. 
Krippendorff's $\alpha$ \cite{alpha} generalises $\kappa$ to 
 data types such as ordinal, interval and ratio data for multiple raters.
It further allows
for distances or weights between categories. 

\citet{related_softirr} calculate soft \gls{IRR} using a variety of distance functions,
but they do not evaluate against a ground truth. 
Furthermore, they only consider BERTScore \cite{bertscore}, BLEU \cite{papineni-etal-2002-bleu},
GLEU \cite{gleu} and the Levenshtein \cite{Levenshtein1965BinaryCC} edit distance, but not other
potential measures such as not yet introduced \glspl{LLM} 
or \gls{NLI}-based measures.
They additionally introduce two novel \gls{IRR} measures based on comparing the distributions of observed and chance
agreement instead of the means.
For simplicity we use the well-known mean-based measures 
in our experiments.
Systematic evaluations of text-based distances for soft \gls{IRR} against a ground truth have never been
conducted to our knowledge.
A recent review also does not discuss soft \gls{IRR} for \gls{NLP} (except edit distances) \cite{nlp_irr_review},
pointing to a research gap.

\section{Methods}

We consider two cases of soft \gls{IRR}: (1) single annotation \gls{IRR} where annotation strings are compared directly and (2) set-matching \gls{IRR} where
two sets of annotations need to be compared. 

\subsection{Soft Inter-Rater Reliability}

Let $A_i$ and $B_i$ denote the annotations of annotator 1 and annotator 2 for document $i$, for $i = 1, \ldots, N$. We
define soft \gls{IRR} measures by replacing the binary agreement function with a soft equivalence measure $\hat{P}(\cdot
\equiv \cdot)$ (Equations~\ref{eq:soft_ao}--\ref{eq:pe_mc}), which can be any
of the previously described equivalence measures. Note that the average agreement is only interpretable to the extent that
the used equivalence measure is interpretable. Cosine similarity, for example, simply reflects the angle between two embeddings, where a given value does not correspond to something with a probabilistic interpretation. Cohen's $\kappa$ just
states how much better the observed agreement is than chance expected agreement, which is interpretable for any measure.

\begin{equation}
p_o^s = \frac{1}{N} \sum_{i=1}^{N} \hat{P}(A_i \equiv B_i)
\label{eq:soft_ao}
\end{equation}

\paragraph{Soft Cohen's $\bm{\kappa}$.}

Soft Cohen's $\kappa$ requires $O(N + K^2)$ time to compute the expected chance agreement $p_e^s$
(Equation~\ref{eq:soft_pe}), where
$N$ is the number of rated examples and $K$ is the number of unique labels.
In the standard case where we have a limited number of labels, this is not too costly due to a small $K$.
With soft \gls{IRR} we will have many more labels, making this
$O(N + N^2)$ in the worst case (all labels occur only once). With computationally intensive measures,
this becomes costly.

We define the expected chance agreement $p_e^s$ as the expected equivalence between two annotations $a \sim
\hat{Q}_1$ and $b \sim \hat{Q}_2$ drawn from the empirical distributions of both annotators
(Equation~\ref{eq:soft_pe}).

\begin{equation}
p_e^s = \mathbb{E}_{a \sim \hat{Q}_1,\; b \sim \hat{Q}_2}\bigl[\hat{P}(a \equiv b)\bigr]
\label{eq:soft_pe}
\end{equation}

We estimate $p_e^s$ using Monte Carlo sampling (Equation~\ref{eq:pe_mc}) or exact computation if the number of
pairwise comparisons is below 1k. For the Monte Carlo estimate we perform $D$ permutations
$\pi_1,\ldots,\pi_D$ of the second annotator's annotations and compute the mean estimated equivalence.

\begin{equation}
\hat{p}_e^s = \frac{1}{DN} \sum_{d=1}^{D} \sum_{i=1}^{N} \hat{P}(A_i \equiv B_{\pi_d(i)})
\label{eq:pe_mc}
\end{equation}

Soft Cohen's $\kappa^s$ is then defined by substituting
$p_o^s$ and $p_e^s$
into Equation~\ref{eq:kappa}. Note that this is Cohen's weighted $\kappa$ \cite{cohen_weighted}, but weights are decided
by a semantic equivalence measure and categories are not predefined.

\paragraph{Soft Fleiss' $\bm{\kappa}$.}
Analogously, we define the soft version of Fleiss' $\kappa$ by replacing the indicator function with the equivalence
estimate (Appendix~\S\ref{sec:appendix_soft_fleiss}, Equations~\ref{eq:soft_fleiss_pi}--\ref{eq:soft_fleiss_pe}).
As for the soft Cohen's $\kappa$, we can estimate the expected chance agreement with Monte Carlo sampling. Note that this
is essentially Krippendorff's $\alpha$ \cite{alpha} with the custom distance $d = 1 - \hat{P}(\cdot \equiv \cdot)$, but
enumerating pairs for $P_e$ with replacement.
For consistent notation, we refer to it
as soft Fleiss' $\kappa^s_{F}$.

\subsection{Set-Assignment}

As mentioned previously, if we consider annotations to be sets, we are presented with an assignment problem between two
sets (Figure~\ref{fig:assignment_toy_example}).
When annotations are sets, let $A = \{a_1, \ldots, a_m\}$ and $B = \{b_1, \ldots, b_n\}$ denote the annotations of the
two annotators for a single document. The set-level equivalence (Equation~\ref{eq:eqv_set}) sums the edge-level
equivalences over the optimal assignment $\phi^*$ (Equation~\ref{eq:hungarian}) and normalises it to the harmonic mean
(see derivation in Appendix~\S\ref{sec:hm_derivation}).

\begin{equation}
\hat{P}_\text{set}(A \equiv B) = \frac{2}{|A| + |B|} \sum_{a \in A} \hat{P}\bigl(a \equiv \phi^*(a)\bigr)
\label{eq:eqv_set}
\end{equation}

Edge weights used for calculating the assignment cost can be obtained with any measure previously described, e.g., edit-based, \gls{NLI}, embeddings or \gls{LLM} equivalence.
We briefly tested the difference between Hungarian and greedy assignment for the set-based experiments, but found no noticeable difference.
However, for the subsequent experiments we use the Hungarian algorithm, since it provides guarantees for optimal assignment.

\section{Experiments}

We initially conduct synthetic experiments to evaluate the performance of different equivalence
approaches with controlled terminologies of known synonyms. In order to show that the results translate
to real-world settings we then conduct experiments on two biomedical datasets
that provide known ground-truth labels as well as free-text annotations.
Unfortunately, data used in our experiments cannot be provided with our code\footnote{\coderepo} and must be obtained
 from the source due to licensing constraints.

\begin{figure*}[t]
    \centering
    \begin{subfigure}[t]{0.440\textwidth}
        \centering
        \includegraphics[width=\linewidth]{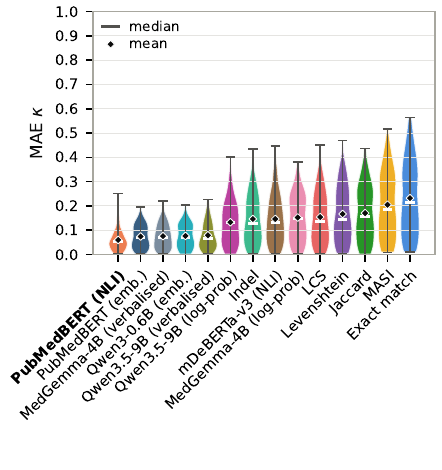}
        \caption{One label per rater.}
        \label{fig:violin_kappa_single}
    \end{subfigure}
    \hfill
    \begin{subfigure}[t]{0.530\textwidth}
        \centering
        \includegraphics[width=\linewidth]{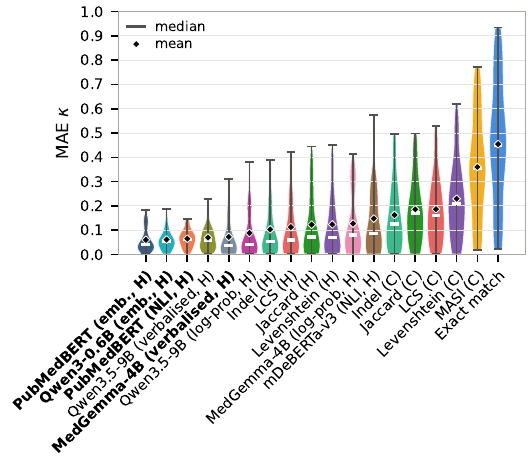}
        \caption{Label set per rater.}
        \label{fig:violin_kappa_set}
    \end{subfigure}
    \caption{Violin plots showing the distribution of the \gls{MAE} of Cohen's $\kappa$ over all synthetic populations 
	(pooled over \acrshort{ICD}-11, \acrshort{MeSH}, \acrshort{MedDRA} and \acrshort{NCIt}) per equivalence measure.
	Measures are ordered by their mean \gls{MAE} of Cohen's $\kappa$.
	Bold labels indicate no significant difference to the best measure according to a paired permutation test with
	10k samples and a significance level of 0.05.
	In the label-set experiments, (C) denotes concatenation of set elements and (H) denotes Hungarian matching.
	}
    \label{fig:synthetic_violins_kappa}
\end{figure*}

\subsection{Synthetic Experiments}
Controlled vocabularies are common in biomedical applications and provide a
natural setting for evaluating soft \gls{IRR} measures. \gls{ICD} \cite{who_icd11} is used to code, for example,
health conditions, diseases and causes of death \cite{icdpaper}. It lists a set of index terms 
per code, which are often alternative spellings or synonyms. \gls{MeSH} \cite{mesh2026} is a controlled
vocabulary for indexing biomedical literature and includes a set of entry terms per concept 
which are alternative words, near-synonyms or closely related concepts. 
\gls{MedDRA} \citepalias{meddra2026} is a terminology for coding adverse events and other medical information in
clinical research and pharmacovigilance
and includes a set of low-level terms (usually synonyms or alternative spellings) per preferred term.

Since the introduced terminologies consist mainly of short-form concepts, we additionally evaluate on the \gls{NCIt},
which is a reference ontology widely used, e.g., for clinical care and translational and basic research related
to cancer \cite{ncit2026}. In contrast to the other terminologies, it contains multiple definitions, typically 1--3 sentences long, from
different contributing sources for a single term. We apply \gls{NCIt} in the same setting as other terminologies, but
sample from the longer-form definitions to generate alternative annotations for a single concept. We use a clean subset
of definitions. For example, we drop templated definitions, remove near duplicates and restrict sources for definitions
(see Appendix~\S\ref{sec:appendix_ncit_filtering}).

We perform 
experiments with all four terminologies by sampling populations
of pairs of terms from the terminologies and randomly replacing all terms with one of their 
synonyms to simulate natural language variation. This is loosely the same procedure applied by \citet{cst}
for corpus shuffling, but applied to paraphrases instead of spans of text. 
We further introduce negatives by replacing
a controlled fraction of the annotations with different concepts. 

\paragraph{Single-label pair generation.} We sample 30 populations of 100 concepts from the terminology. For each
concept $c$, we form a pair $(a, b)$ by replacing its preferred name with two uniformly sampled synonyms.
We further sample a target agreement level between 0 and 1 per population, denoting the fraction of pairs that should
be positives (agreement), and randomly replace the concepts for the second annotator ($b$) with a different concept for the
remaining pairs. We then compute the ground-truth \gls{IRR} measures from the
unchanged pairs of initially sampled concepts (before replacing with synonyms) using exact matching of the concepts.
We perform $D = 3$ permutations to calculate the expected chance agreement $p_e^s$.

\paragraph{Set-level pair generation.} For the set case we sample 15 populations of 100 pairs of annotation sets and use the Jaccard coefficient before synonym replacement to calculate the ground-truth \gls{IRR} measures.
Sampled sets contain between 2 and 5 concepts, where the set-size difference is dependent on the sampled agreement level.
We then replace each concept in the sets with a synonym and randomly replace a fraction of the concepts in the second set with negatives dependent on the size mismatch and target agreement level.
We perform $D = 2$ permutations to calculate $p_e^s$.

\paragraph{Evaluation.} We test whether different equivalence measures can sufficiently approximate the ground-truth \gls{IRR} by measuring the \gls{MAE} between the estimated soft Cohen's $\kappa$ and the ground-truth Cohen's $\kappa$.
We provide additional results for Scott's $\pi$ in the appendix, since we observed that the performance on these two 
measures is similar. 

\paragraph{Models.} For each deep-learning-based equivalence estimate, we evaluate two models: one smaller specialist model trained for the biomedical domain and one larger generalist model. In order to enable reproducibility, we exclusively evaluate open-weights models. Specifically, we choose PubMedBERT based models as specialists for embedding and \gls{NLI}-models. The \gls{NLI} variant \cite{pubmedbert_nli} is a PubMedBERT fine-tuned on MedNLI \cite{mednli}, whereas the embedding model \cite{neuml_emb} was fine-tuned on PubMed title-abstract pairs. As a specialist \gls{LLM} we rely on MedGemma 1.5 4B \cite{medgemma}, which is based on a Gemma 3 model post-trained on biomedical datasets. 
As generalists, we use Qwen3-Embedding-0.6B \cite{qwen3} for embeddings and mDeBERTa-v3 for \gls{NLI}, which was
fine-tuned on over 2.7M premise-hypothesis pairs \cite{gennli}. For a generalist \gls{LLM} we rely on Qwen3.5-9B
\cite{qwen35}. We disable reasoning capabilities in all \gls{LLM} experiments to enable us to calculate Monte Carlo
estimates of chance agreement in feasible time.

\paragraph{Results.}
Figure~\ref{fig:synthetic_violins_kappa} shows the distribution of the \gls{MAE} of Cohen's $\kappa$ over the different
synthetic populations of the datasets.
We see a clear advantage of the best deep-learning-based measures (emb., \gls{NLI}, \gls{LLM}) over the exact, set and edit measures.
However, the weaker deep learning measures are matched or beaten by edit-based measures,
and in the set setting the bipartite edit-based and set-overlap measures remain competitive. Note that the
exact measure might still recover the right distance when both annotators happen to sample the same synonym by chance.

\begin{figure*}[t]
    \centering
    \includegraphics[width=\textwidth]{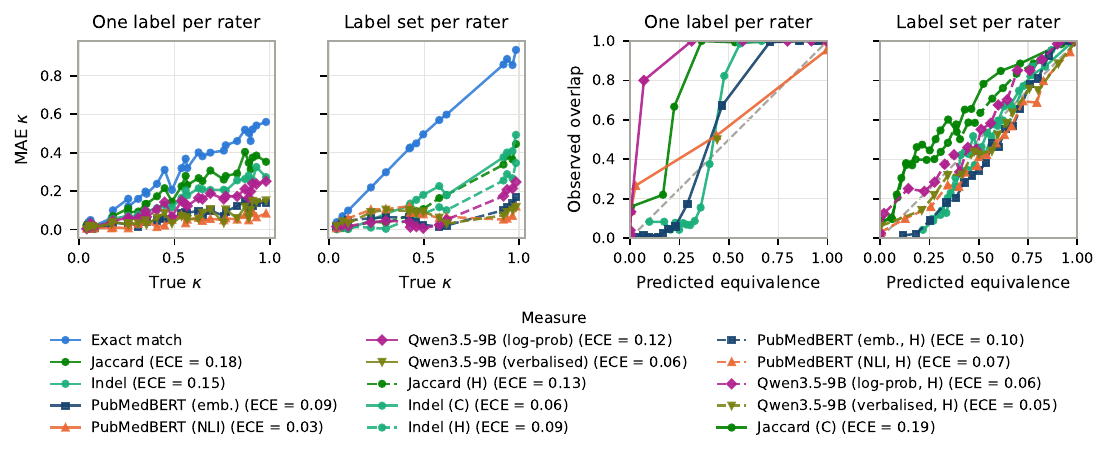}
	\caption{Plots using \acrshort{MeSH} synthetic experiment data showing the \gls{MAE} on $\kappa$ as a function of the
		ground truth and calibration plots
		for the different equivalence estimates in both the single-label and set configuration.
	The \gls{ECE} of each measure is shown in its legend entry. Exact matching and the best model per
	group (set-based, edit-based, emb., \acrshort{NLI}, \acrshort{LLM} log-prob, \acrshort{LLM} verbalised) are shown.
	In the label-set experiments, (C) denotes concatenation of set elements and (H) denotes Hungarian matching.
}
    \label{fig:synthetic_mesh_overview}
\end{figure*}

Figure~\ref{fig:synthetic_mesh_overview} displays the \gls{MAE} of the soft Cohen's $\kappa$
as a function of the ground-truth $\kappa$ for both the set and single-label settings. 
We observe that virtually all measures have a harder time accurately measuring high agreement than low agreement. 
This observation is even more pronounced for the set experiment, which is expected,
since an exact match is even more unlikely in the set case.
However, the rate at which the \gls{MAE} increases is considerably lower for all soft measures compared to exact.

\subsection{Real-World Experiments}
We conduct experiments on two real-world datasets that have unstructured text annotations
and known ground-truth mappings to standardised terminologies. Such datasets are unfortunately very
rare, or only consider span-level annotations as opposed to free-text generated by annotators.

\paragraph{Derm1M \cite{derm1m}.}
Derm1M 
contains
roughly 1M image-text pairs of dermatology images and free-text captions gathered from various online sources, 
including YouTube, PubMed articles, medical forums and public datasets.
It has been found that it contains a large number of duplicate
images with different free-text annotations \cite{skinmap}. For each image, the authors additionally provide a mapping to a standardised dermatology terminology for diagnoses and skin concepts, which can
be used as a ground truth for evaluating soft \gls{IRR} measures. We combine both skin concepts and diagnoses into a single ground-truth set for each image. 
Filtering images by the following criteria, we end up with 336 test images: (1) at least two captions, (2) annotations are mapped to the terminology (diseases, skin concepts), and (3) annotations are from different data sources.
Unfortunately, the dataset does not provide any information about annotators, so we simply treat each data source as a separate annotator.


\paragraph{\acrshort{REFLACX} \cite{reflacx}.} The \gls{REFLACX} dataset was constructed with the goal of providing implicit
localisation data by eye tracking of radiologists while reading chest X-rays. It additionally includes free-text annotations for each X-ray by up to five
radiologists for phases 1 and 2 of the data collection (109 X-rays). Additionally, each annotator selected all suitable
labels from a closed set of anomalies, which we use as ground truth.

\paragraph{Evaluation.} We again calculate the ground-truth \gls{IRR} measures by exact matching or Jaccard on the standardised labels
and then compare them to the estimated soft \gls{IRR} measures from the free-text annotations. In contrast to the synthetic experiments, we additionally report Fleiss' $\kappa$ as a measure, since we have more than two annotators per example in these datasets.
Cohen's $\kappa$ is the pairwise average of all annotator pairs.

\paragraph{Results.} The results for the real-world datasets are shown in Table~\ref{tab:real-mae},
with the estimates underlying the reported errors in Appendix Table~\ref{tab:real-all-full}.
The best soft measures achieve \glspl{MAE} between 0.005 and 0.065 for Cohen's or Fleiss' $\kappa$, which
is accurate enough to warrant their use in practice. While the best measure to use changes depending on the dataset, 
we observe that the best performing measures are generally the deep-learning-based soft measures, with PubMedBERT
(\gls{NLI}) and Qwen3.5-9B (verbalised) among the best measures in three of the four settings. For completeness, we also report additional \gls{IRR} statistics
introduced by \citet{related_softirr} in Appendix Table~\ref{tab:real-all-full}.

Note that this evaluation is limited, since it only considers the subdomains dermatology and radiology.
However, the restrictive requirements on dataset properties did not allow us to find public data for more subdomains.
A dataset must provide multiple annotators per example, free-text annotations not verbatim from the example, a
categorical ground-truth label and anonymous annotator IDs.
Furthermore, it is unclear how the result would translate to agreement on longer annotations, such as complete
paragraphs, reports or letters, since we only look at image captions or annotations that are typically only a few
sentences long.
We would also expect longer-form texts to be harder to measure agreement on, since they may contain several
claims, which makes partial agreement more likely.

\begin{table}[t]
\centering
\small
\resizebox{\linewidth}{!}{
\begin{tabular}{@{}lrrrr@{}}
\toprule
 & \multicolumn{2}{c}{\textbf{Derm1M}} & \multicolumn{2}{c}{\textbf{REFLACX}} \\
\cmidrule(lr){2-3} \cmidrule(lr){4-5}
\textbf{Measure} & $\bm{\kappa}$ & $\bm{\kappa_F}$ & $\bm{\kappa}$ & $\bm{\kappa_F}$ \\
\midrule
  Exact match & 0.174 & 0.221 & 0.375 & 0.374 \\
  Jaccard & \textbf{0.065} & \textbf{0.073} & 0.344 & 0.351 \\
  MASI & 0.123 & 0.148 & 0.367 & 0.368 \\
  Levenshtein & \textbf{0.074} & 0.106 & 0.368 & 0.375 \\
  Indel & \textbf{0.060} & 0.086 & 0.360 & 0.367 \\
  LCS & \textbf{0.090} & 0.117 & 0.355 & 0.361 \\
\midrule
  PubMedBERT (emb.) & \textbf{0.081} & 0.112 & 0.208 & 0.223 \\
  Qwen3-0.6B (emb.) & \textbf{0.053} & 0.105 & 0.197 & 0.209 \\
\midrule
  PubMedBERT (NLI) & 0.097 & \textbf{0.037} & \textbf{0.094} & \textbf{0.093} \\
  mDeBERTa-v3 (NLI) & \textbf{0.072} & 0.089 & 0.282 & 0.267 \\
\midrule
  Qwen3.5-9B (log-prob) & 0.076 & 0.116 & 0.344 & 0.338 \\
  MedGemma-4B (log-prob) & \textbf{0.005} & \textbf{0.072} & 0.258 & 0.241 \\
\midrule
  Qwen3.5-9B (verbalised) & \textbf{0.076} & 0.085 & \textbf{0.061} & \textbf{0.065} \\
  MedGemma-4B (verbalised) & 0.139 & 0.236 & 0.129 & 0.122 \\
\bottomrule
\end{tabular}

}
\caption{Results on the real-world datasets showing the \gls{MAE} against the ground-truth \glspl{IRR} calculated by
the Jaccard coefficient on the standardised label sets. All methods not significantly different from the best
according to a paired permutation test (significance level 0.05) with 10k samples are bold.}
\label{tab:real-mae}
\end{table}


\section{Error Analysis}

\begin{figure*}[t]
    \centering
    \includegraphics[width=0.905\textwidth]{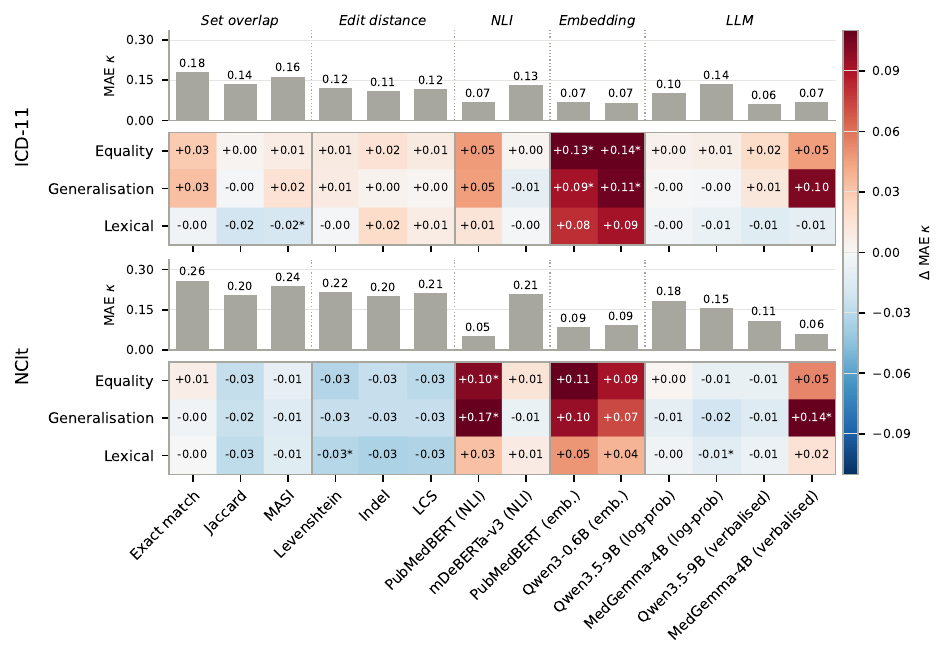}
	\caption{
		$\Delta$ \gls{MAE} against the random-negative baseline for Cohen's $\kappa$ for different negative sampling
		conditions and equivalence measures on \gls{ICD}-11 and \gls{NCIt}.
		Asterisks ($^*$) indicate a significant difference to the baseline according to a Holm-corrected permutation test with 10k samples.
	}
    \label{fig:neg_heatmap}
\end{figure*}

Using standardised terminologies allows us to systematically study the failure modes of different
equivalence measures by controlling the type of disagreement between annotators. We do this by changing the
way we sample negative examples in the synthetic
experiments. We first apply the sampling mechanism on preferred terms in the terminology (not synonyms) and
then again randomly choose synonyms to simulate free-text annotations.
We investigate three different types of errors:

\begin{itemize}
\setlength\itemsep{-0.3em}
	\item \textbf{Equality Errors:} The annotators mention similar concepts that are not semantically equivalent.
		To investigate this, we sample siblings in the hierarchy of the terminology as
		negatives.
	\item \textbf{Generalisation Errors:} The annotators mention closely related concepts, where one is more
		general than the other and the two are therefore not equivalent.
		Negatives are sampled using parents or grandparents (or their synonyms) of a concept from the terminology.
	\item \textbf{Lexical Errors:} The annotators mention concepts that are semantically distinct, but share a
		similar lexical form, e.g., overlapping strings.
		We sample negatives by normalised Indel distance (10 closest synonyms of other concepts).
\end{itemize}

Figure~\ref{fig:neg_heatmap} shows the change in mean \gls{MAE} of Cohen's $\kappa$ against the random-negative baseline for
the different negative sampling conditions on \gls{ICD}-11 and \gls{NCIt}.
For most measures on \gls{ICD}-11, the different negative sampling conditions 
do not significantly affect the \gls{MAE}. However, for embedding-based
approaches, 
all of the conditions increase the error considerably. We hypothesise that embedding-based approaches do not
differentiate between similarity and equivalence, and thus are not able to distinguish between similar but
non-equivalent annotations. However, on longer-form annotations, such as \gls{NCIt} definitions, the PubMedBERT
\gls{NLI} model suffers from a similar issue. \gls{LLM}-based log-prob variants (Equation~\ref{eq:ent_dir}), mDeBERTa-v3 (\gls{NLI}) and
Qwen3.5-9B (verbalised) are the most stable, changing the \gls{MAE} by at most 0.02. However, only Qwen3.5-9B
(verbalised) also reaches low error rates.

Because of this, embedding-based approaches might be replaced with \gls{NLI}-based ones for short-form
annotations, as a cheaper alternative to \gls{LLM}-based ones, and with \gls{LLM}-based ones for longer-form
annotations.
However, \gls{LLM}-based methods require substantially more compute even without reasoning due to the
Monte Carlo estimate of the expected
agreement by chance (see Appendix Table~\ref{tab:distance-properties}). \gls{NLI} based measures can still work well on longer form texts, but seem to be more sensitive to the tested error types.
The full per-dataset error distributions behind these means are given in Appendix~\S\ref{sec:appendix_negative_sampling}. 

\section{Conclusion}
We revisit the problem of measuring \gls{IRR} for complex annotation tasks where annotators provide unstructured text.
Using controlled experiments on synthetic datasets generated using biomedical terminologies, we compare various semantic equivalence measures and show how well they approximate Cohen's $\kappa$.
We perform experiments on real-world radiology and dermatology datasets, where both structured and
free-text annotations are available.
To our knowledge, this is the first time soft \gls{IRR} estimates have been validated against ground-truth agreement
for unstructured text annotations.

We find that the choice of measure has a significant impact on the estimated \gls{IRR}, and that
different measures have different failure modes.
For example, embedding-based measures are the most scalable and perform well on randomly sampled negatives, but their
error increases the most when close negatives are sampled.
\gls{LLM}-based measures using log-probabilities are the most stable under close negatives, but they do not reach the
lowest errors and are roughly three orders of magnitude slower than embeddings, which makes the estimate of chance
agreement expensive.

We therefore suggest \gls{NLI}-based measures as a compromise for short-form annotations, since they are roughly 50 to
100 times faster than \gls{LLM}-based measures, and PubMedBERT (\gls{NLI}) is among the best measures in three of the
four real-world settings.
For longer-form annotations, such as definitions, \gls{LLM}-based measures may be preferable if the compute is
available.
However, we only compare two models per type, so more experiments are needed to confirm this.

\clearpage

\section*{Limitations}
\begin{itemize}
	\item The study is limited to biomedical datasets, which may not generalise to other domains.
	\item Our set of tested equivalence measures is not exhaustive, and other measures may yield different results.
	\item Synthetic experiments may not fully capture the real world, since the terminologies used are likely very common in pretraining corpora and thus may present easier cases than real-world annotations.
	\item We deliberately did not consider larger models, since calculating $p_e^s$ with them would not be feasible in practice
	due to the cost of the Monte Carlo estimate (consider Appendix Table~\ref{tab:distance-properties}).
    \item We do not explicitly show that our soft-\gls{IRR} measures correlate with downstream model performance.
	Previous work has shown that, with standard \gls{IRR}, such correlations exist but they are not a ceiling
	\cite{boguslav2017interannotator,richie-etal-2022-inter}.
	This only transfers completely to soft \gls{IRR} if the \gls{MAE} is uncorrelated with the ground-truth $\kappa$,
	as otherwise the error affects this relationship.
	This is not the case, since we found that high $\kappa$ values are harder to approximate in our experiments.
\end{itemize}

\section*{Acknowledgements}
This research is supported through computing resources by the ADAPT
Centre for Digital Content Technology, which is funded under the SFI Research Centres Programme (Grant 13/RC/2106\_P2) and is co-funded under the European Regional Development Fund (ERDF). One author was supported by the Swiss National Science Foundation (SNSF) under grant 20HW-1\_228541.
The authors thank the reviewers for their insightful and helpful comments. 

We used Claude Code to assist in the following under close supervision: setting up experiments, plotting results,
reviewing the paper, rephrasing and identifying potentially overlooked literature.

\bibliography{custom}

\appendix
\clearpage

\section{Methods Supplement}

\subsection{Fleiss' \texorpdfstring{$\bm{\kappa}$}{κ}}
\label{sec:appendix_fleiss}

To calculate Fleiss' $\kappa$, we first compute the extent of agreement $P_i$ among the $R$ raters for the $i^{\text{th}}$ example
(Equation~\ref{eq:fleiss_pi}).

\begin{equation}
P_i := \frac{1}{R(R-1)} \sum_{r=1}^{R} \sum_{\substack{r'=1 \\ r' \neq r}}^{R}
    \mathbf{1}\bigl[l_i^{(r)} = l_i^{(r')}\bigr] \label{eq:fleiss_pi}
\end{equation}

The average of $P_i$ over all examples gives the observed agreement $P_o$ (Equation~\ref{eq:fleiss_po}).

\begin{equation}
P_o := \frac{1}{N} \sum_{i=1}^{N} P_i \label{eq:fleiss_po}
\end{equation}

To calculate the chance agreement $P_e$, we first compute the empirical label distribution $\hat{p}_k$ over all $NR$
annotations (Equation~\ref{eq:fleiss_emp_pe}).

\begin{equation}
\hat{p}_k = \frac{1}{NR} \sum_{i=1}^{N} \sum_{r=1}^{R} \mathbf{1}\bigl[l_i^{(r)} = k\bigr] \label{eq:fleiss_emp_pe}
\end{equation}

To produce $P_e$ we sum the squared empirical label frequencies (Equation~\ref{eq:fleiss_pe}).

\begin{equation}
P_e := \sum_{k} \hat{p}_k^{\,2} \label{eq:fleiss_pe}
\end{equation}

Fleiss' $\kappa$ is then calculated like Cohen's (Equation~\ref{eq:fleiss_kappa}).

\begin{equation}
\kappa_F := \frac{P_o - P_e}{1 - P_e} \label{eq:fleiss_kappa}
\end{equation}

\subsection{Soft Fleiss' \texorpdfstring{$\bm{\kappa}$}{κ}}
\label{sec:appendix_soft_fleiss}

\begin{align}
P_i^s &= \frac{1}{R(R-1)} \sum_{r=1}^{R} \sum_{\substack{r'=1 \\ r' \neq r}}^{R}
    \hat{P}\bigl(A_i^{(r)} \equiv A_i^{(r')}\bigr) \label{eq:soft_fleiss_pi}\\
P_e^s &= \mathbb{E}_{a,\,b \sim \hat{Q}}\bigl[\hat{P}(a \equiv b)\bigr] \label{eq:soft_fleiss_pe}
\end{align}

Here, $A_i^{(r)}$ is the annotation of rater $r$ for example $i$ and $\hat{Q}$ is the empirical distribution of all
$NR$ annotations. $P_o^s$ and $\kappa_F^s$ follow from Equations~\ref{eq:fleiss_po} and~\ref{eq:fleiss_kappa} with
$P_i^s$ and $P_e^s$ in place of $P_i$ and $P_e$.

\subsection{Scott's \texorpdfstring{$\bm{\pi}$}{π}}

\begin{figure*}[t]
    \centering
    \begin{subfigure}[t]{0.442\textwidth}
        \centering
        \includegraphics[width=\linewidth]{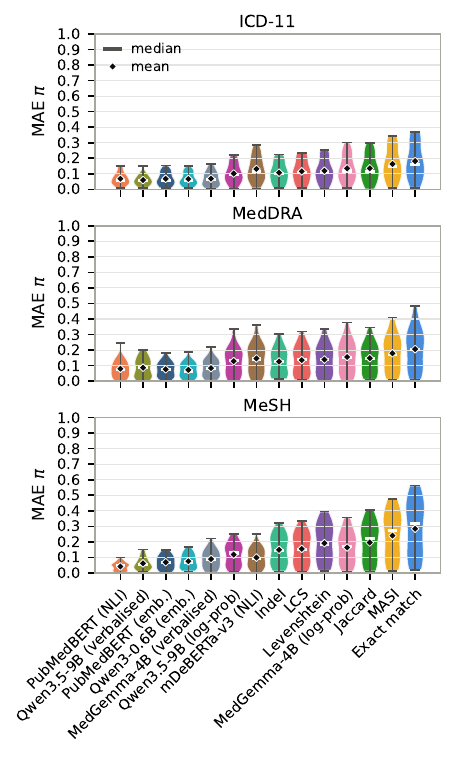}
        \caption{One label per rater.}
        \label{fig:violin_pi_single}
    \end{subfigure}
    \hfill
    \begin{subfigure}[t]{0.528\textwidth}
        \centering
        \includegraphics[width=\linewidth]{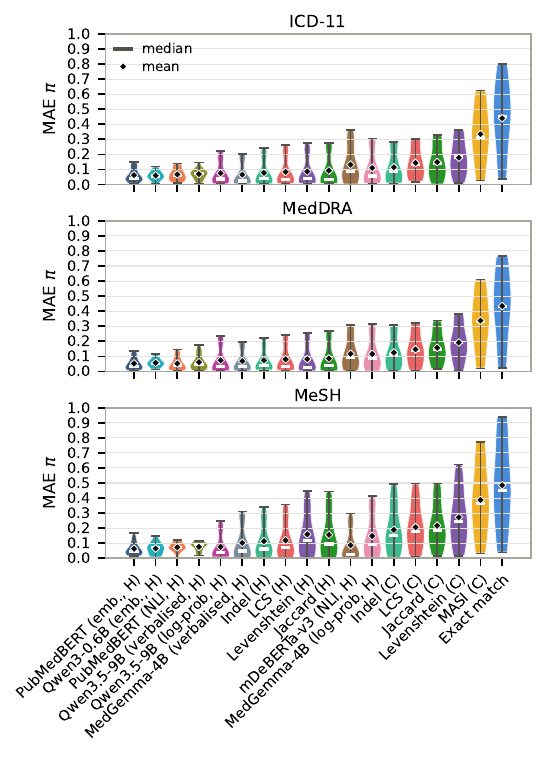}
        \caption{Label set per rater.}
        \label{fig:violin_pi_set}
    \end{subfigure}
    \caption{Violin plots showing the distribution of the \gls{MAE} of Scott's $\pi$ over all synthetic populations
	separated by terminology (\gls{ICD}-11, \gls{MedDRA} and \gls{MeSH}) and per equivalence measure.
	Measures are ordered by their mean \gls{MAE} of Scott's $\pi$ pooled over the three terminologies.
	In the label-set experiments, (C) denotes concatenation of set elements and (H) denotes Hungarian matching.}
    \label{fig:synthetic_violins_pi}
\end{figure*}

Scott's $\pi$ differs from Cohen's $\kappa$ in that it assumes that annotators share the same empirical label distribution, which Cohen's $\kappa$ does not.
This means under Scott's $\pi$ we do not expect annotators to have different individual biases in how often they choose a label.
Concretely, we calculate the expected chance agreement $p_e^\pi$ using the average label frequencies
across annotators $\bar{p}_k$ (Equations~\ref{eq:scotts_pi1}--\ref{eq:scotts_pi2}).
Figure~\ref{fig:synthetic_violins_pi} shows the corresponding synthetic results for Scott's $\pi$.

\begin{align}
\bar{p}_k &= \tfrac{1}{2}\!\left(\hat{p}_k^{(1)} + \hat{p}_k^{(2)}\right) \label{eq:scotts_pi1}
\\
p_e^\pi &:= \sum_{k} \bar{p}_k^{2} \label{eq:scotts_pi2}
\end{align}

\subsection{Edit-Distance Equivalence Measures}
The underlying distances describe how many characters need to be changed to transform one string into another. In the
Levenshtein equivalence, all operations (insertion, deletion, substitution) have a cost of 1, while substitution has a
cost of 2 in the Indel equivalence. Normalisation then divides the Levenshtein distance by the maximum of the two string
lengths and the Indel distance by their sum (Equations~\ref{eq:norm_esim}--\ref{eq:indel}). The \gls{LCS} equivalence
(Equation~\ref{eq:lcs}) that we use is based on the same longest common subsequence as the Indel equivalence, but
normalised by the maximum string length.

\begin{align}
\hat{P}_\text{Lev}(a \equiv b) &= 1 - \tfrac{\text{Lev}(a,b)}{\max(|a|,|b|)} \label{eq:norm_esim} \\
\hat{P}_\text{Indel}(a \equiv b) &= 1 - \tfrac{\text{Indel}(a,b)}{|a|+|b|} \label{eq:indel}\\
\hat{P}_\text{LCS}(a \equiv b) &= \tfrac{|\text{LCS}(a,b)|}{\max(|a|,|b|)} \label{eq:lcs}
\end{align}

\subsection{Set-Overlap Equivalence Measures}
The Jaccard equivalence is defined by Equation~\ref{eq:jaccard} and the \gls{MASI} equivalence is defined by
Equation~\ref{eq:masi}. The \gls{MASI} equivalence is a weighted version of the Jaccard equivalence that penalises
non-subset relations between two sets, where the weighting $M(A,B)$ is defined as in Equation~\ref{eq:masi_weighting}.

\begin{equation}
    M(A,B) = \begin{cases}
        1 & \text{if } A = B \\
        \tfrac{2}{3} & \text{if } A \subsetneq B \text{ or } B \subsetneq A \\
        \tfrac{1}{3} & \text{if } A \cap B \neq \emptyset \text{ (non-subset)} \\
        0 & \text{if } A \cap B = \emptyset
    \end{cases}
	\label{eq:masi_weighting}
\end{equation}

\begin{equation}
    J(A, B) = \frac{|A \cap B|}{|A \cup B|}
	\label{eq:jaccard}
\end{equation}

\begin{equation}
\label{eq:masi}
\text{MASI}(A, B) = J(A, B) \cdot M(A, B)
\end{equation}

\subsection{Harmonic-Mean Derivation}
\label{sec:hm_derivation}

Let $\Sigma$ be the sum of matched edge scores as in

\begin{equation*}
    \Sigma = \sum_{a \in A} \hat{P}\left(a \equiv \phi^*(a)\right).
\end{equation*}

Let precision be $p = \frac{\Sigma}{|A|}$ and recall be $r  = \frac{\Sigma}{|B|}$. This is the total score divided by the number of elements in the first set (precision) or by the number of elements in the second set (recall).
We know that the $F_1$ score is the harmonic mean of precision and recall and that it can be written as

\begin{equation*}
    F_1 = \frac{2}{\frac{1}{r} + \frac{1}{p}}
\end{equation*}

Then we get Equation~\ref{eq:eqv_set} as follows.

\begin{align*}
F_1 &= \frac{2}{\frac{|A|}{\Sigma} + \frac{|B|}{\Sigma}} \\
&= \frac{2}{\frac{|A|+|B|}{\Sigma}}\\
&= 2 \cdot \frac{\Sigma}{|A|+|B|} \\
&= \frac{2}{|A|+|B|} \cdot \Sigma
\end{align*}\medskip

\section{Experiments Supplement}

\subsection{NCIt Definition Filtering}
\label{sec:appendix_ncit_filtering}
Before using the \gls{NCIt} definitions in our experiments, we filter them to improve the quality of the dataset we
test on.

We keep only active concepts and drop retired or obsolete ones.
We remove concepts containing ``CTCAE'' in the name, which are grading criteria that are largely copies of a base
concept.
Concepts that use the role R88 (Disease Is Stage), and all their descendants, are dropped as well, since they are
largely copies of a base concept with different stages.
We further remove concepts with the semantic type \textit{Intellectual Product} (2,126 concepts, or 16\% of the
concepts that would otherwise remain), since in a sample we inspected they were mostly questionnaire items, answer
options, trial documents and code lists of data standards rather than biomedical concepts.
We also drop concepts with the semantic type \textit{Laboratory Procedure}, since their definitions are largely
templated and not free-text.
Concepts with an executive order notice in any of their definitions are removed too.

Furthermore, we keep only definitions from an allowlist of 23 definition sources and remove definitions stating that
they might be outdated or that contain copyright text.
We clean the remaining definitions by cutting field labels (``EXAMPLE(S):'', ``NOTE(S):'' and ``OTHER NAME(S):'')
as well as notes and cross-references (``NOTE:'', ``See also'', ``Contrast with'' and ``Compare to''), each together
with the text that follows them.
We also remove citations in square brackets, URLs, trailing years preceded by a double hyphen and trailing source tags.
We then remove definitions with fewer than 5 tokens, near-duplicate definitions of the same concept (token Jaccard
similarity of at least 0.7) and templated definitions, whose text with the concept name masked occurs at least 10 times
for the same source.

Finally, we keep only concepts that have at least 2 definitions after filtering, retaining 10,992 concepts with 23,751
definitions.
The three most frequent definition sources are NCI, CDISC and NCI-GLOSS.
More details on the filtering can be found in our published code\footnote{\coderepo}.

\subsection{Synthetic Set-Pair Generation}

We generate synthetic set pairs $(A, B)$, such that the expected Jaccard coefficient $\mathbb{E}[J(A, B)]$ is approximately equal to a target agreement level $\alpha \in [0, 1]$.
We first sample the set size $|A|$, and then sample $|B|$, 
such that the size difference $\delta = |B| - |A|$ is bounded by a gap $g_\alpha$ that closes as $\alpha$ increases.
The gap $g_\alpha$ is the minimum of two terms: a population-level budget $\lfloor (1-\alpha)(s_{\max} - s_{\min}) \rceil$,
where $s_{\min}$ and $s_{\max}$ bound the set size, and a per-pair budget $\lfloor (1-\alpha)|A| \rfloor$ that keeps the
difference small for small $|A|$. We then draw $\delta$ uniformly from $\{-g_\alpha, \ldots, g_\alpha\}$ and clip
$|B| = |A| + \delta$ to $[s_{\min}, s_{\max}]$.
Furthermore, $k^{*}$ is the expected number of shared concepts between $A$ and $B$ that would yield a Jaccard coefficient of $\alpha$.
The final number of shared concepts $k$ is then the random rounding of $k^{*}$ to one of the two nearest integers, capped at $\min(|A|, |B|)$.

\begin{figure*}[t]
    \centering
    \begin{subfigure}[t]{0.442\textwidth}
        \centering
        \includegraphics[width=\linewidth]{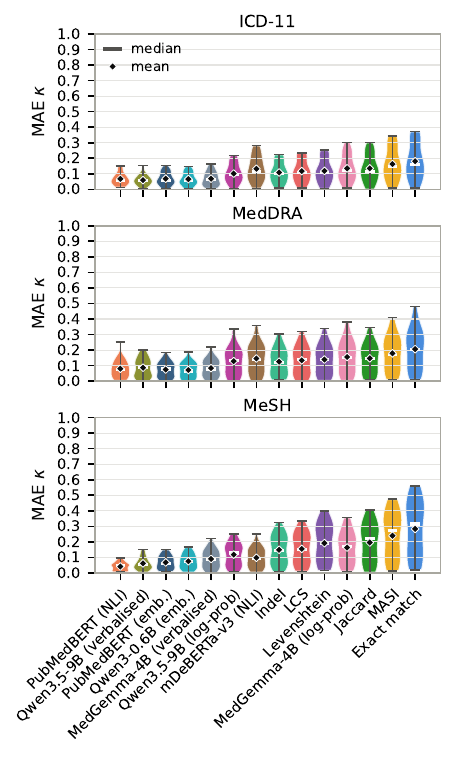}
        \caption{One label per rater.}
        \label{fig:violin_kappa_single_perdataset}
    \end{subfigure}
    \hfill
    \begin{subfigure}[t]{0.528\textwidth}
        \centering
        \includegraphics[width=\linewidth]{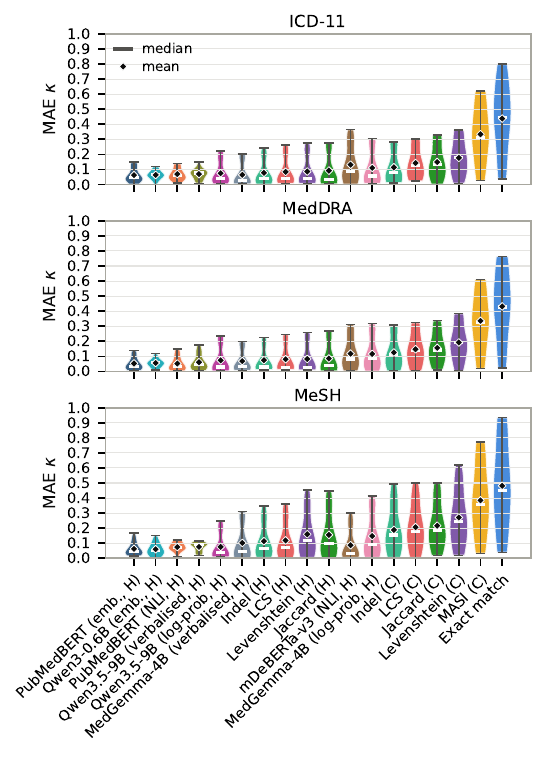}
        \caption{Label set per rater.}
        \label{fig:violin_kappa_set_perdataset}
    \end{subfigure}
    \caption{Violin plots showing the distribution of the \gls{MAE} of Cohen's $\kappa$ over all synthetic populations
	separated by terminology (\gls{ICD}-11, \gls{MedDRA} and \gls{MeSH}) and per equivalence measure.
	Measures are ordered by their mean \gls{MAE} of Cohen's $\kappa$ pooled over the three terminologies.
	In the label-set experiments, (C) denotes concatenation of set elements and (H) denotes Hungarian matching.}
    \label{fig:synthetic_violins_kappa_perdataset}
\end{figure*}

\begin{figure*}[t]
    \centering
    \begin{subfigure}[t]{0.442\textwidth}
        \centering
        \includegraphics[width=\linewidth]{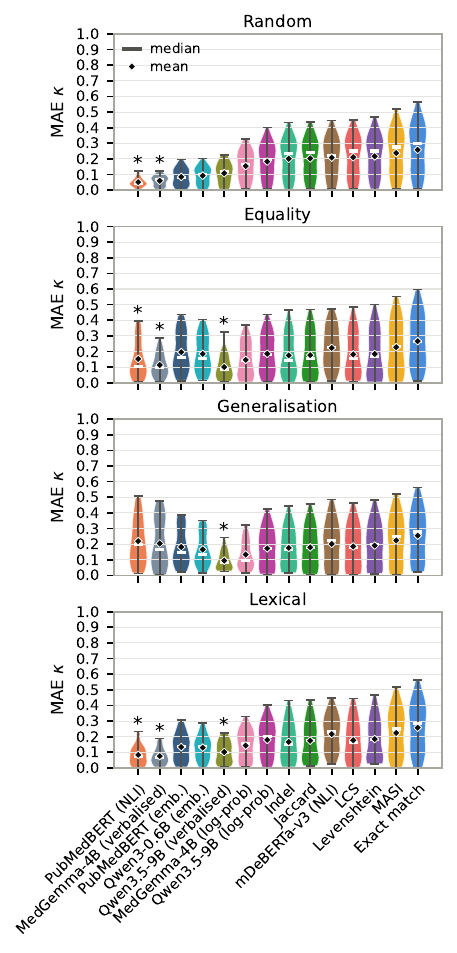}
        \caption{One label per rater.}
        \label{fig:violin_ncit_kappa_single}
    \end{subfigure}
    \hfill
    \begin{subfigure}[t]{0.528\textwidth}
        \centering
        \includegraphics[width=\linewidth]{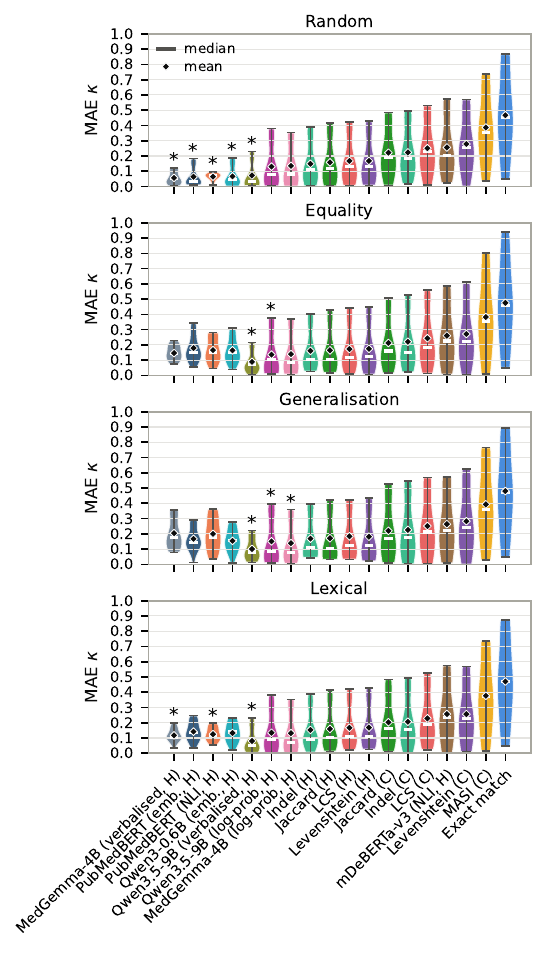}
        \caption{Label set per rater.}
        \label{fig:violin_ncit_kappa_set}
    \end{subfigure}
    \caption{Violin plots showing the distribution of the \gls{MAE} of Cohen's $\kappa$ over all synthetic populations
	of \gls{NCIt} separated by negative sampling condition and per equivalence measure.
	Measures are ordered by their mean \gls{MAE} of Cohen's $\kappa$ under random negatives.
	Asterisks ($^*$) mark the measures whose mean \gls{MAE} is not significantly different from the best
	measure in the same panel according to a paired permutation test with 10k samples and a significance level of 0.05.
	In the label-set experiments, (C) denotes concatenation of set elements and (H) denotes Hungarian matching.}
    \label{fig:synthetic_violins_ncit_kappa}
\end{figure*}

\subsection{Negative Sampling Conditions}
\label{sec:appendix_negative_sampling}

\begin{figure*}[t]
    \centering
    \includegraphics[width=\textwidth]{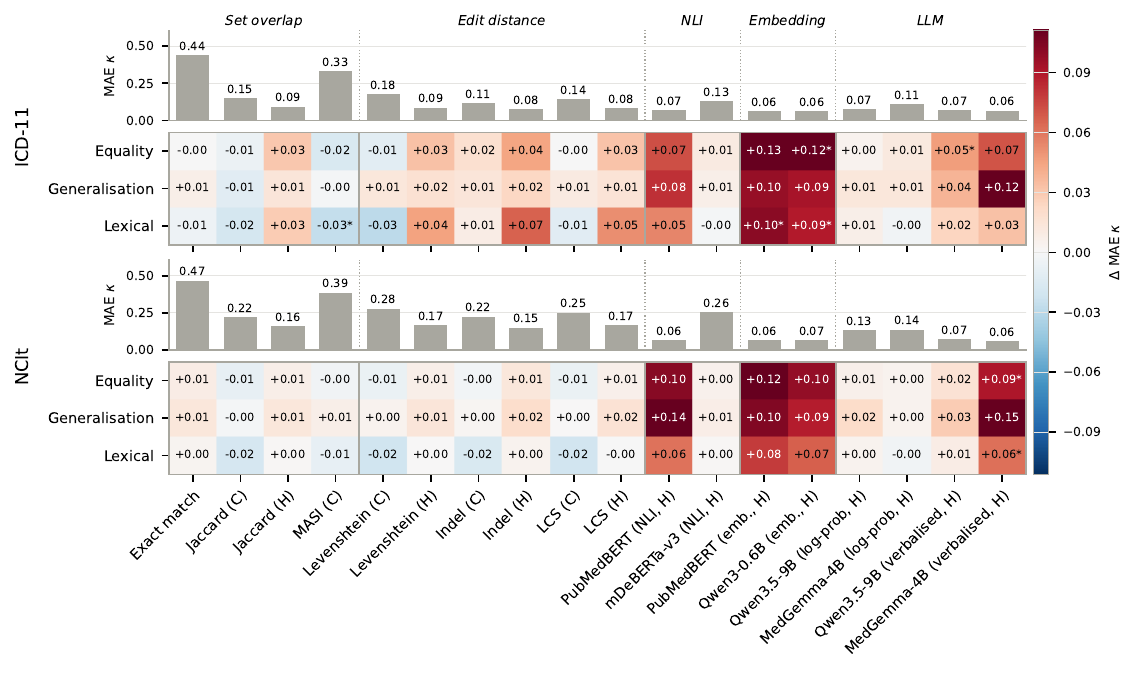}
	\caption{
     $\Delta$ \gls{MAE} against the random-negative baseline for Cohen's $\kappa$ for different negative sampling
	 conditions and equivalence measures on the set-valued annotations of \gls{ICD}-11 and \gls{NCIt}.
	 Asterisks ($^*$) indicate a significant difference to the baseline according to a Holm-corrected permutation test with 10k samples.
	 In the label-set experiments, (C) denotes concatenation of set elements and (H) denotes Hungarian matching.
	}
    \label{fig:neg_heatmap_set}
\end{figure*}

Figures~\ref{fig:violin_neg_sibling_kappa}--\ref{fig:violin_neg_lexical_kappa} show the full distributions behind the
\gls{ICD}-11 heatmaps in Figures~\ref{fig:neg_heatmap} and~\ref{fig:neg_heatmap_set}: the \gls{MAE} of Cohen's
$\kappa$ per equivalence measure for the three harder negative-sampling conditions, with the single-label setting on
the left and the label-set setting on the right. The random-negative baseline is shown in
Figure~\ref{fig:synthetic_violins_kappa_perdataset}.
Measures
follow the same ordering in every panel, namely by mean \gls{MAE} of Cohen's $\kappa$ under random negatives.
The corresponding results on \gls{NCIt} are shown in Figures~\ref{fig:synthetic_violins_ncit_kappa}
and~\ref{fig:neg_heatmap_set}.

\newcommand{\negviolinpair}[1]{%
    \begin{subfigure}[t]{0.442\textwidth}
        \centering
        \includegraphics[width=\linewidth]{figures/synthetic_violin_neg_single_#1_kappa_stacked.pdf}
        \caption{One label per rater.}
    \end{subfigure}
    \hfill
    \begin{subfigure}[t]{0.528\textwidth}
        \centering
        \includegraphics[width=\linewidth]{figures/synthetic_violin_neg_set_#1_kappa_stacked.pdf}
        \caption{Label set per rater.}
    \end{subfigure}%
}

\begin{figure*}[t]
    \centering
    \negviolinpair{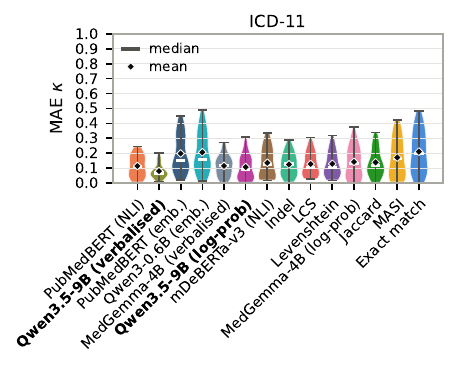}
	\caption{Violin plots showing the distribution of the \gls{MAE} of Cohen's $\kappa$ on \gls{ICD}-11 for the
	\textbf{equality} (siblings) negative sampling condition.
	Bold labels indicate no significant difference to the best measure according to a paired permutation test with
	10k samples and a significance level of 0.05.
	In the label-set experiments, (C) denotes concatenation of set elements and (H) denotes Hungarian matching.}
    \label{fig:violin_neg_sibling_kappa}
\end{figure*}

\begin{figure*}[t]
    \centering
    \negviolinpair{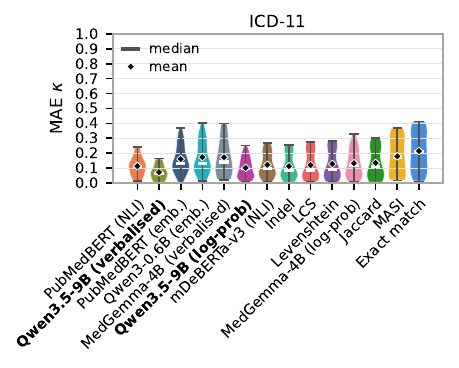}
	\caption{Violin plots showing the distribution of the \gls{MAE} of Cohen's $\kappa$ on \gls{ICD}-11 for the
	\textbf{generalisation} (ancestors) negative sampling condition.
	Bold labels indicate no significant difference to the best measure according to a paired permutation test with
	10k samples and a significance level of 0.05.
	In the label-set experiments, (C) denotes concatenation of set elements and (H) denotes Hungarian matching.}
    \label{fig:violin_neg_parent_kappa}
\end{figure*}

\begin{figure*}[t]
    \centering
    \negviolinpair{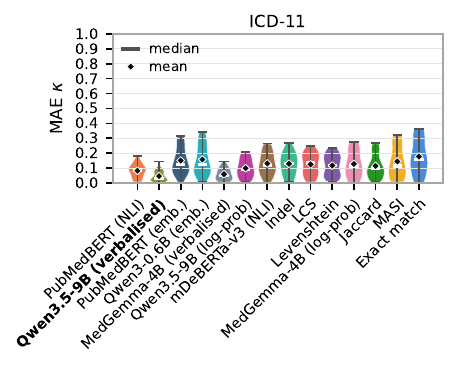}
	\caption{Violin plots showing the distribution of the \gls{MAE} of Cohen's $\kappa$ on \gls{ICD}-11 for the
	\textbf{lexical} (surface-form overlap) negative sampling condition.
	Bold labels indicate no significant difference to the best measure according to a paired permutation test with
	10k samples and a significance level of 0.05.
	In the label-set experiments, (C) denotes concatenation of set elements and (H) denotes Hungarian matching.}
    \label{fig:violin_neg_lexical_kappa}
\end{figure*}

\subsection{Soft Reliability Across Terminologies}

\begin{figure*}[t]
    \centering
    \includegraphics[width=\textwidth]{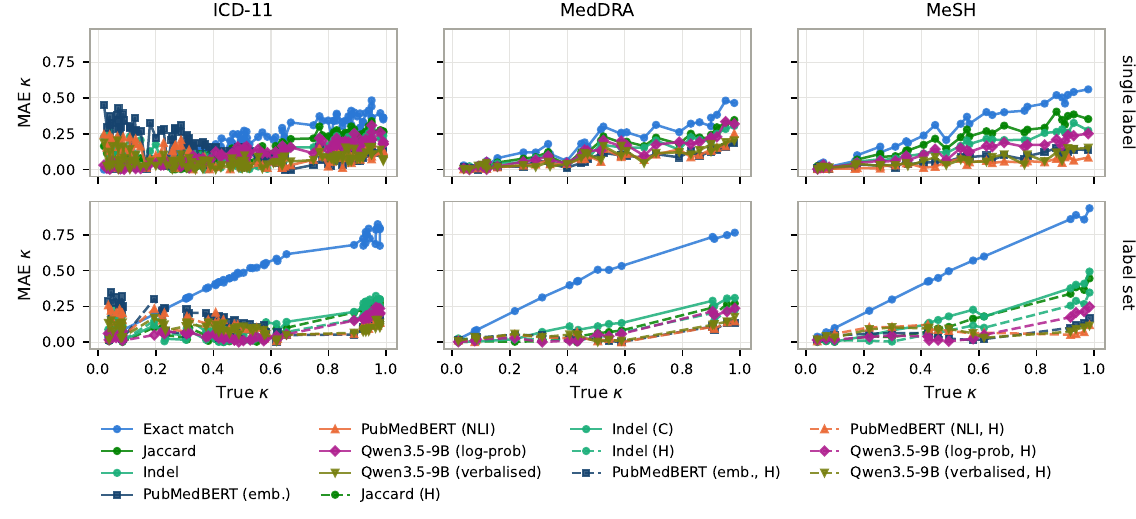}
	\caption{Plots showing the \gls{MAE} on $\kappa$ as a function of the
		ground truth for the different equivalence estimates in both the single-label and set configuration for
		\gls{ICD}-11, \gls{MedDRA} and \gls{MeSH} separately.
	Exact matching and the best model per
	group (set-based, edit-based, emb., \acrshort{NLI}, \acrshort{LLM} log-prob, \acrshort{LLM} verbalised) are shown.
	In the label-set experiments, (C) denotes concatenation of set elements and (H) denotes Hungarian matching.
}
    \label{fig:synthetic_mae_kappa_vs_true_full}
\end{figure*}

Figure~\ref{fig:synthetic_mesh_overview} in the main body shows the \gls{MeSH} panels only.
Figure~\ref{fig:synthetic_mae_kappa_vs_true_full} gives the same curves for \gls{ICD}-11, \gls{MedDRA} and
\gls{MeSH}.

\subsection{Computational Cost}
\label{sec:appendix_distance_properties}

Table~\ref{tab:distance-properties} shows the throughput, GPU hours and number of prompts of each deep-learning-based
equivalence measure.
\begin{table}[t]
\centering
\small
\resizebox{\columnwidth}{!}{
\begin{tabular}{lrrr}
\toprule
\textbf{Measure} & \textbf{Pairs/s} & \textbf{GPU (h)} & \textbf{Prompts} \\
\midrule
PubMedBERT (emb.) & 1,077 & <0.01 & 23,632 \\
Qwen3-0.6B (emb.) & 1,117 & <0.01 & 23,632 \\
PubMedBERT (NLI) & 103 & 0.10 & 749,091 \\
mDeBERTa-v3 (NLI) & 80 & 0.12 & 749,091 \\
Qwen3.5-9B (log-prob) & 1.66 & 5.95 & 746,734 \\
MedGemma-4B (log-prob) & 1.33 & 7.43 & 746,734 \\
Qwen3.5-9B (verbalised) & 1.02 & 9.68 & 430,888 \\
MedGemma-4B (verbalised) & 0.97 & 10.21 & 430,888 \\
\bottomrule
\end{tabular}

}
\caption{Computational cost of the set experiments using random uniform negative sampling,
summed over \acrshort{ICD}-11, \acrshort{MedDRA} and \acrshort{MeSH} (15 populations of 100 set-valued
annotation pairs each) on an NVIDIA RTX A6000 GPU. Pairs/s is the mean over the three experiments.}
\label{tab:distance-properties}
\end{table}

\subsection{Estimates on the Real-World Datasets}

Table~\ref{tab:real-mae} in the main body reports the \gls{MAE} of every measure against the
ground-truth \gls{IRR}. Table~\ref{tab:real-all-full} additionally lists the agreement each
measure actually estimates, next to the ground-truth value it is compared against.

\begin{table*}[t]
\centering
\small
\resizebox{\textwidth}{!}{
\begin{tabular}{lrrrrrrrrrrrrrrrr}
\toprule
 & \multicolumn{8}{c}{\textbf{Derm1M}} & \multicolumn{8}{c}{\textbf{REFLACX}} \\
\cmidrule(lr){2-9} \cmidrule(lr){10-17}
 & \multicolumn{2}{c}{$\bm{\kappa}$} & \multicolumn{2}{c}{$\bm{\kappa_F}$} & \multicolumn{2}{c}{$\bm{\sigma}$} & \multicolumn{2}{c}{\textbf{KS}} & \multicolumn{2}{c}{$\bm{\kappa}$} & \multicolumn{2}{c}{$\bm{\kappa_F}$} & \multicolumn{2}{c}{$\bm{\sigma}$} & \multicolumn{2}{c}{\textbf{KS}} \\
\cmidrule(lr){2-3} \cmidrule(lr){4-5} \cmidrule(lr){6-7} \cmidrule(lr){8-9} \cmidrule(lr){10-11} \cmidrule(lr){12-13} \cmidrule(lr){14-15} \cmidrule(lr){16-17}
\textbf{Measure} & \textbf{est.} & \textbf{MAE} & \textbf{est.} & \textbf{MAE} & \textbf{est.} & \textbf{MAE} & \textbf{est.} & \textbf{MAE} & \textbf{est.} & \textbf{MAE} & \textbf{est.} & \textbf{MAE} & \textbf{est.} & \textbf{MAE} & \textbf{est.} & \textbf{MAE} \\
\midrule
  \textit{Ground truth} & 0.191 & -- & 0.272 & -- & 0.422 & -- & 0.388 & -- & 0.374 & -- & 0.374 & -- & 0.372 & -- & 0.451 & -- \\
\midrule
  Exact match & 0.017 & 0.174 & 0.051 & 0.221 & 0.049 & 0.373 & 0.049 & 0.339 & -0.001 & 0.375 & 0.000 & 0.374 & 0.000 & 0.372 & 0.000 & 0.451 \\
  Jaccard & \textbf{0.126} & \textbf{0.065} & \textbf{0.199} & \textbf{0.073} & \textbf{0.387} & \textbf{0.035} & \textbf{0.359} & \textbf{0.029} & 0.030 & 0.344 & 0.023 & 0.351 & 0.049 & 0.322 & 0.159 & 0.292 \\
  MASI & 0.068 & 0.123 & 0.124 & 0.148 & \textbf{0.425} & \textbf{0.003} & \textbf{0.388} & \textbf{0.000} & 0.007 & 0.367 & 0.006 & 0.368 & 0.049 & 0.322 & 0.159 & 0.292 \\
  Levenshtein & \textbf{0.117} & \textbf{0.074} & 0.166 & 0.106 & 0.306 & 0.116 & 0.284 & 0.104 & 0.006 & 0.368 & -0.001 & 0.375 & 0.019 & 0.353 & 0.059 & 0.392 \\
  Indel & \textbf{0.131} & \textbf{0.060} & 0.186 & 0.086 & 0.318 & 0.104 & 0.306 & 0.082 & 0.015 & 0.360 & 0.007 & 0.367 & 0.025 & 0.347 & 0.058 & 0.393 \\
  LCS & \textbf{0.101} & \textbf{0.090} & 0.155 & 0.117 & 0.257 & 0.165 & 0.229 & 0.159 & 0.019 & 0.355 & 0.013 & 0.361 & 0.044 & 0.327 & 0.077 & 0.374 \\
\midrule
  PubMedBERT (emb.) & \textbf{0.271} & \textbf{0.081} & 0.384 & 0.112 & \textbf{0.413} & \textbf{0.009} & 0.479 & 0.091 & 0.166 & 0.208 & 0.151 & 0.223 & 0.072 & 0.299 & 0.236 & 0.215 \\
  Qwen3-0.6B (emb.) & \textbf{0.244} & \textbf{0.053} & 0.377 & 0.105 & \textbf{0.460} & \textbf{0.038} & 0.484 & 0.096 & 0.177 & 0.197 & 0.166 & 0.209 & 0.123 & 0.249 & 0.285 & 0.166 \\
\midrule
  PubMedBERT (NLI) & 0.288 & 0.097 & \textbf{0.309} & \textbf{0.037} & \textbf{0.355} & \textbf{0.066} & 0.465 & 0.077 & \textbf{0.280} & \textbf{0.094} & \textbf{0.281} & \textbf{0.093} & \textbf{0.355} & \textbf{0.017} & 0.256 & 0.195 \\
  mDeBERTa-v3 (NLI) & \textbf{0.119} & \textbf{0.072} & 0.183 & 0.089 & 0.272 & 0.150 & 0.267 & 0.121 & 0.092 & 0.282 & 0.108 & 0.267 & 0.118 & 0.254 & 0.219 & 0.231 \\
\midrule
  Qwen3.5-9B (log-prob) & 0.114 & 0.076 & 0.156 & 0.116 & 0.503 & 0.081 & 0.609 & 0.221 & 0.030 & 0.344 & 0.036 & 0.338 & 0.126 & 0.245 & \textbf{0.467} & \textbf{0.016} \\
  MedGemma-4B (log-prob) & \textbf{0.186} & \textbf{0.005} & \textbf{0.200} & \textbf{0.072} & 0.517 & 0.095 & 0.559 & 0.171 & 0.116 & 0.258 & 0.133 & 0.241 & \textbf{0.375} & \textbf{0.003} & \textbf{0.487} & \textbf{0.036} \\
\midrule
  Qwen3.5-9B (verbalised) & \textbf{0.114} & \textbf{0.076} & 0.187 & 0.085 & 0.263 & 0.159 & 0.243 & 0.145 & \textbf{0.313} & \textbf{0.061} & \textbf{0.310} & \textbf{0.065} & \textbf{0.286} & \textbf{0.086} & \textbf{0.392} & \textbf{0.058} \\
  MedGemma-4B (verbalised) & 0.330 & 0.139 & 0.508 & 0.236 & \textbf{0.396} & \textbf{0.026} & 0.549 & 0.161 & 0.503 & 0.129 & 0.496 & 0.122 & 0.097 & 0.275 & 0.358 & 0.093 \\
\bottomrule
\end{tabular}
}
\caption{Results on the real-world datasets showing the \gls{MAE} against the ground-truth \glspl{IRR} calculated by
the Jaccard coefficient on the standardised label sets. All methods not significantly different from the best
according to a paired permutation test (significance level 0.05) with 10k samples are bold.
We also show the ground-truth agreement and the estimate by each measure here. Furthermore, we show $\sigma$
and \gls{KS} as additional \gls{IRR} measures
introduced by \citet{related_softirr}. They operate on comparing the distributions of $p_o$ and $p_e$ instead of the
means like $\kappa$ statistics do.}
\label{tab:real-all-full}
\end{table*}

\end{document}